\documentclass[sigconf,screen=true,review=false,authorversion]{acmart}
\copyrightyear{2024}
\acmYear{2024}
\setcopyright{acmlicensed}\acmConference[SIGSIM PADS '24]{38th ACM SIGSIM Conference on Principles of Advanced Discrete Simulation}{June 24--26, 2024}{Atlanta, GA, USA}
\acmBooktitle{38th ACM SIGSIM Conference on Principles of Advanced Discrete Simulation (SIGSIM PADS '24), June 24--26, 2024, Atlanta, GA, USA}
\acmDOI{10.1145/3615979.3656058}
\acmISBN{979-8-4007-0363-8/24/06}

\acmSubmissionID{3965}
\acmPrice{}

\begin{document}
\title[Towards Learning Stochastic Population Models by Gradient Descent]{Towards Learning Stochastic Population Models\\by Gradient Descent}

\author{Justin N.\ Kreikemeyer}
\orcid{0000-0002-4109-3608}
\email{justin.kreikemeyer@uni-rostock.de}
\affiliation{%
  \institution{Institute for Visual and Analytic Computing, University of Rostock}
  \streetaddress{Albert-Einstein-Straße 22}
  \city{Rostock}
  \country{Germany}
  \postcode{18059}
}
\author{Philipp Andelfinger}
\orcid{0000-0002-0211-7136}
\email{philipp.andelfinger@uni-rostock.de}
\affiliation{%
  \institution{Institute for Visual and Analytic Computing, University of Rostock}
  \streetaddress{Albert-Einstein-Straße 22}
  \city{Rostock}
  \country{Germany}
  \postcode{18059}
}
\author{Adelinde M.\ Uhrmacher}
\orcid{0000-0001-5256-4682}
\email{adelinde.uhrmacher@uni-rostock.de}
\affiliation{%
  \institution{Institute for Visual and Analytic Computing, University of Rostock}
  \streetaddress{Albert-Einstein-Straße 22}
  \city{Rostock}
  \country{Germany}
  \postcode{18059}
}

\begin{abstract} 
Increasing effort is put into the development of methods for learning mechanistic models from data. 
This task entails not only the accurate estimation of parameters but also a suitable model structure.
Recent work on the discovery of dynamical systems formulates this problem as a linear equation system.
Here, we explore several simulation-based optimization approaches, which allow much greater freedom in the objective formulation and weaker conditions on the available data.
We show that even for relatively small stochastic population models, simultaneous estimation of parameters and structure poses major challenges for optimization procedures.
Particularly, we investigate the application of the local stochastic gradient descent method, commonly used for training machine learning models.
We demonstrate accurate estimation of models but find that enforcing the inference of parsimonious, interpretable models drastically increases the difficulty.
We give an outlook on how this challenge can be overcome.
\end{abstract}

\ccsdesc[300]{Computing methodologies~Discrete-event simulation}
\ccsdesc[500]{Computing methodologies~Modeling methodologies}
\ccsdesc[100]{Computing methodologies~Machine learning}

\keywords{automatic model generation, gradient descent, stochastic simulation algorithm, discrete-event simulation, differentiable simulation}

\maketitle

\section{Introduction}
\label{sec:introduction}

Statistical machine learning methods provide exciting advances in automatically learning (deep) models from data.
Whereas these models exhibit impressive predictive abilities \cite{noe2020machine}, their black-box nature does not directly contribute to understanding the reference system's mechanics and impedes precise manual refinement. 
This motivated the development of methods for automatically deriving (white-box) mechanistic models from data \cite{Nobile2013reverse,Brunton2016discover,Klimovskaia2016sparseR,Burrage2024usingA,martinelli2023reactmine}.
With these, manual, hypothesis-driven knowledge discovery can increasingly be augmented by automatic, data-driven approaches \cite{maass2018data}. 
Such an automatic modeling approach is useful when (parts of) the mechanisms of the reference system are unknown, but there are measurements of its behavior over time.
Learning mechanistic models from data then entails not only parameter estimation but also the \emph{simultaneous} identification of a suitable model structure. 

In this paper, we study 
learning stochastic, \emph{discrete-event} models with an underlying continuous representation of time from \emph{time-series} snapshots of some traversed state distributions by gradient descent. 
Specifically, we focus on Markovian \emph{population models} that are expressed as reaction systems. 
Our contributions are: 
\begin{itemize}
    \item \autoref{sec:learning_reaction_systems_with_gradient_descent} provides different possible formulations of the model learning problem.
    \item \autoref{sec:reparametrization} shows how reparametrization enables parameter estimation over different orders of magnitude.
    \item \autoref{sec:discussion_of_results} provides first results on the simultaneous learning of structure and parameters by gradient descent. It discusses the challenges and opportunities of the approach. 
\end{itemize}

\noindent We briefly introduce the reaction system formalism in \autoref{sec:population_based_modeling} and stochastic gradient estimation in \autoref{sec:stochastic_gradient_estimation}.
\autoref{sec:related_work} reviews related work.
After presenting our methods in \autoref{sec:learning_reaction_systems_with_gradient_descent} as outlined above, we conclude in \autoref{sec:discussion_of_results}.

\section{Population-based Modeling}
\label{sec:population_based_modeling}

In the biology and chemistry domains, \emph{reaction systems} are a commonly used modeling formalism \cite{keating2020sbml}.
They describe system dynamics in terms of the consumption and production of entities at certain rates. 
Their underlying assumption is that entities can be grouped into homogeneous populations of \emph{species} 
$S_i, i\in\{1,\dots,n_S\}$ 
residing in a \emph{well-stirred} medium.
A reaction takes the form 
\vspace{-0.05cm}
$$R_i:\sum_{j=1}^{n_S}c_{ij}S_j\xrightarrow{r_i}\sum_{j=n_S+1}^{2n_S}c_{ij}S_{j-n_S}$$
\vspace{-0.15cm}

\noindent with $\mathbf{C}\in\mathbb{N}^{n_R\times2n_S}$ being a matrix of \emph{coefficients} (``model structure''), $\mathbf{r}$ the vector of \emph{rate constants} (``parameters''), and $n_R, n_S$ the number of reactions and species in the system, respectively. 
A reaction system can be completely represented by providing $\mathbf{C}$ and $\mathbf{r}$. 
A vector of species counts gives the starting conditions of a reaction system, i.e., $\mathbf{S}_{init}$.

As a running example, consider the well-studied SIR model of disease spread, comprising three species representing populations of \textbf{s}usceptible, \textbf{i}nfected and \textbf{r}ecovered individuals:

\noindent\begin{tabular}{ccc}
  $\begin{aligned}
   & R_0\colon 1S + 1I & \xrightarrow{0.02} & \text{ }\fcolorbox{red}{white}{2}I\\
   & R_1\colon 1I & \xrightarrow{5.00} & \text{ }1R
  \end{aligned}$
  &
  $\Rightarrow\kern-5pt$
  &
  $\begin{aligned}
    \mathbf{C}_{SIR} &= \left(
    \begin{array}{cccccc}
         1 & 1 & 0 & 0 & \fcolorbox{red}{white}{2} & 0  \\
         0 & 1 & 0 & 0 & 0 & 1 
    \end{array}
    \right)\\
    \mathbf{r}_{SIR} &= \left(
    \begin{array}{cc}
         0.02 & 5.00
    \end{array}
    \right)^T
  \end{aligned}$
\end{tabular}

\noindent This reaction system has two reactions with coefficient matrix $\mathbf{C}_{SIR}$ and rate vector $\mathbf{r}_{SIR}$.
The first reaction describes the infection of a susceptible individual and the second its recovery.
Note the correspondence between entries in $\mathbf{C}$ and $R_i$ indicated by the red box.
Species participating with coefficient 0 are omitted from $R_i$.
We will use $\mathbf{S}_{init}=\left(1980\,\,20\,\,0\right)$ as the initial state for the S, I, and R species, respectively.

Population-based models defined as reaction systems can be simulated either through \emph{numerical integration} with ordinary differential equation (ODE) semantics \cite{kurtz1972relationship,hahl2016comparison}
or the \emph{stochastic simulation algorithm} (SSA) \cite{Gillespie1976aGenera} with continuous-time Markov chain (CTMC) semantics.  
In many cases, stochastic effects cannot be ignored  \cite{Ramaswamy2012discrete,Mcadams1999itsAN}. 
Therefore, 
instead of focusing on the mean continuous dynamics, our approach will take the stochasticity of the system into account. 

The vector of species counts $\mathbf{S}_t$ fully represents the state of the model at the current time $t$.
We make the common assumption that the transition probabilities are governed by the probability of two entities in the well-stirred medium reacting, so the transitions of the CTMC are governed by the stochastic mass action law \cite{kurtz1972relationship,Gillespie1976aGenera}. 
The effective rate of a reaction in a given state is called its \emph{propensity} $\alpha$.
For example, for the SIR model, we have $\alpha_0=0.02 \cdot S \cdot I$, i.e., the more susceptible and infected individuals there are, the likelier an infection event is to happen.
Note that other functions may be used to calculate the propensity depending on the modeled system.
Another common assumption is that the probability of more than two species colliding (interacting) is very low.
Thus, we only consider binary reactions with at most two reactants.
Despite making these assumptions here for simplicity, our approach is theoretically able to accommodate any dependence of the propensities on the state as well as n-ary reactions.

As a simulator, we use Gillespie's direct method \cite{Gillespie1976aGenera}, which takes sample trajectories through the CTMC defined by $\mathbf{C}$ and $\mathbf{r}$ using a Monte Carlo strategy.
At each event, $t$ is advanced according to an exponential distribution based on the sum of the propensities $\alpha_i$.
The state is updated by choosing from a categorical distribution over the reactions, subtracting the reactants, and adding the products.
With the number of samples approaching infinity,
the probability distribution over system states and time (likelihood) is obtained.

\section{Stochastic Gradient Estimation}
\label{sec:stochastic_gradient_estimation}

When there is a closed form of the likelihood, its gradient is an effective tool for optimization.
However, a closed form is unattainable for many real-world systems, necessitating Gillespie's SSA.
Determining the gradient of this algorithm is not straightforward.
The well-established method of automatic differentiation (AD) provides performant means to calculate the gradient of algorithms at runtime \cite{Margossian2019aReview}.
However, this gradient cannot account for the jumps (discontinuities) inherent to individual SSA trajectories, resulting from the discrete state changes.
So even with the mean over trajectories being a smooth function, AD is not useful for optimization.

Thus, we resort to recent advances in estimating the gradient of an alternative objective function, which is smoothed over jumps \cite{Kreikemeyer2023smoothin}.
We use a finite-differences estimator with stochastic step-size for simplicity, cf.~\cite{Polyak1987introduc} (Chapter 3.4) and for further analysis \cite{Nesterov2017randomG}:
\begin{equation}
 \nabla f(\mathbf{\theta}) \approx \frac{1}{N}\sum_{n=1}^{N} \frac{f(\mathbf{\theta}+\sigma\mathbf{u}) - f(\mathbf{\theta})}{\sigma}\mathbf{u} 
\end{equation}

\noindent where $\mathbf{\theta}$ is the parameter vector, $\sigma$ is a smoothing factor that determines the smoothing applied to the objective $f$, and $\mathbf{u}\sim\mathcal{N}(\mathbf{0},\mathbf{I})$ is a vector of i.i.d.\ normal variates with mean $0$ and variance $1$.
In contrast to finite differences, which need at least one sample per dimension of $\mathbf{\theta}$, through simultaneous perturbation, this estimator requires only two samples for estimating the full gradient.
For the number of samples $n$ approaching infinity, the estimate converges to the gradient of a smoothed version of $f$ \cite{Nesterov2017randomG}.
Further, it can handle jumps and noise in the objective through the smoothing controlled by $\sigma$.

\section{Related Work}
\label{sec:related_work}

Originating from system identification \cite{kozin1986system}, learning mechanistic models has recently inspired various research in many application fields \cite{tan2023automatic,Askari2023evolutio,martinelli2023reactmine}.
Related to our work, two major approaches can be distinguished: genetic programming, which for the first time provided strategic means of searching in the space of programs or models 
\cite{koza2000reverse,Nobile2013reverse}, and sparse regression, which identifies short yet accurate symbolic expressions, such as differential equations \cite{Daniels2015automate,Brunton2016discover}.
Recently, these approaches have also been combined, e.g., to discover multibody physics systems \cite{Askari2023evolutio}.

Specifically in the case of reaction networks, \cite{Nobile2013reverse} proposed genetic programming to identify reaction systems with ODE semantics.
A population of candidate structures is evolved, and evolutionary operators are applied based on the candidates' fitness.
To accurately rank a structure, its fitness is determined by the best solution found with particle swarm optimization and numerical integration.
The authors of \cite{martinelli2023reactmine} propose a statistical search algorithm called Reactmine to infer chemical reactions with ODE semantics.
In \cite{Klimovskaia2016sparseR}, the sparse identification of non-linear dynamics (SINDy) \cite{Brunton2016discover} is adapted to the stochastic semantics (cf. \autoref{sec:population_based_modeling}).
This is achieved by working with the moment-equations of the CTMC, an ODE system describing the time-evolution of the Markov chain's moments.
A two-step regression approach is employed to achieve robustness against heteroscedastic, noisy measurements and reaction constants of different magnitudes.

A recent publication adjusts the SINDy approach to accommodate coupled differential equations such as those resulting from the ODE semantics of reaction networks \cite{Burrage2024usingA};
\cite{jiang2022identification} also brings SINDy to the case of biochemical systems with mass-action kinetics accounting for uncertainty and enabling an informed model selection.

In contrast to the above, here we aim at a simulation-based optimization approach, which also allows, e.g., the straightforward inclusion of unmeasured species, 
arbitrary kinetics, and accounting for probability distributions (instead of their moments).
Further, our proposed methods do not rely on numerical differentiation of the time-series data, which can be inaccurate in the presence of noise and large or uneven sampling intervals.

Using gradient descent for parameter estimation of simulation models also saw great interest recently \cite{andelfinger2023towards,chopra2023differentiable}, including biochemical reaction systems \cite{wang2010parameter}.
In \cite{yang2020bayesian}, gradient descent enables Bayesian inference over general ODE models.

\section{Learning Reaction Systems with Gradient Descent}
\label{sec:learning_reaction_systems_with_gradient_descent}

Consider a reaction system $\mathbf{R}$ with coefficients $\mathbf{C}$, stochastic rate constants $\mathbf{r}\in\mathbb{R}^n$ and initial populations $\mathbf{S}_\emph{init}$.
Assuming the structure $\mathbf{C}$ of the model is known, we can simulate trajectories over states $\mathbf{S}_t, t\geq0$ of the CTMC given by $\mathbf{C}$ and a certain parametrization $\mathbf{r}$.
Typically, we want the trajectories produced by $\mathbf{R}$ to resemble the behavior of a reference system.
To achieve this, suitable parameter values $\mathbf{r}$ have to be estimated from collected time-series data: 
Given measurements $D_t$ at discrete times $t\in\{1,\dots,n\}$, the goal is to maximize the likelihood $\mathcal{L}(D;\mathbf{r})$ or some other measure of goodness of fit. 
Determining the parameters $\mathbf{r}$ that maximize the likelihood is also referred to as the inverse problem, since ``forward'' simulation provides a sample from $\mathcal{L}$ for a given $\mathbf{r}$.

Here, our goal is to simultaneously infer the structure of the model, i.e., we try to find $\mathbf{r}$ \emph{and} $\mathbf{C}$, such that $\mathcal{L}(D;\mathbf{C},\mathbf{r})$ is maximal.
Obviously, this is a much harder task than just estimating parameters, as the degrees of freedom in the inverse problem are drastically increased.
Further, the optimization landscape will exhibit additional jumps, introduced by the discrete entries in $\mathbf{C}$.
In fact, we can formulate the problem with varying degrees of smoothness (prior to considering a smoothed objective, cf.~\autoref{sec:stochastic_gradient_estimation}). 
The following formulations are adapted to the goal of recovering the SIR model (cf. \autoref{sec:population_based_modeling}), which we later use for evaluation.

\emph{Library of Reactions.}
Our first problem formulation is inspired by the use of reaction libraries in \cite{Burrage2024usingA,Klimovskaia2016sparseR}.
This approach can directly be translated to a simulation-based optimization problem: the reaction system to optimize comprises (a selection of) all reactions for a given number of species.
The task is to adjust $\mathbf{r}$, where reactions $i$ with $r_i$ below a certain threshold are dropped from the final model.
Our library consists of the $36$ binary reactions that abide by the conservation law $S+I+R=2000$.
This problem is completely smooth in all dimensions.

\emph{Coefficient Steps.}
In the second problem formulation, we fixate the number of reactions to two and try to adjust $\mathbf{C}$ with $c_{ij}\in\{0,1,2\}$ and $\mathbf{r}$ directly, yielding a $14$-dimensional problem.
This problem is non-smooth in the coefficient dimensions.

\emph{Reaction Steps.}
In the third formulation, we again work with a library of reactions but introduce a (continuous) ranking vector of the same dimensionality as $\mathbf{r}$.
In each simulation run, only the two reactions with the highest rank are considered, enforcing a certain model size.
The task is then to adjust the ranking together with the two rate parameters, one for each reaction in the top two.

\emph{Library of Systems.}
The final formulation, which we adopt for didactic purposes, is a brute-force approach.
It simultaneously optimizes the $1260$ rates for all possible combinations of two reactions from our library of $36$.
With one optimization per model being much more performant, this example showcases the gradient estimator's ability to steer the rate adjustment for large numbers of structures.

Generally, more than one reaction system can produce trajectories from the distribution in $D$ \cite{craciun2008identifiability}. 
It is often hard to choose the ``right'' system automatically, so the choice must involve domain experts \cite{jiang2022identification}. 
However, certain criteria can constrain the optimization process to desirable solutions, such as parsimony (choosing a low number of reactions producing a good fit) and prior knowledge (such as number of species, conservation laws, or even known reactions).
Some of these constraints may result in an NP-hard problem for which the best-known solution is brute force \cite{gupte2022finegrained}.
This can be overcome, e.g., by regularization (like in SINDy) and relaxation.

As we will demonstrate on the example of the problems above, there is a tradeoff between the ability to strongly enforce these constraints and the smoothness of the objective function, which in turn determines the difficulty of the optimization task.

\begin{figure}[b!]
    \vspace{-0.4cm}
    \centering
    \includegraphics[width=\columnwidth]{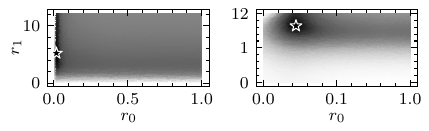}
    \vspace{-0.9cm}
    \caption{The SIR model's response surface (left) and the effect of reparametrization (right). A darker color equals a lower loss and the star marks the optimum.}
    \label{fig:reparametrization}
\end{figure}

\begin{figure*}[!ht]
    \centering
    \includegraphics[width=0.97\textwidth]{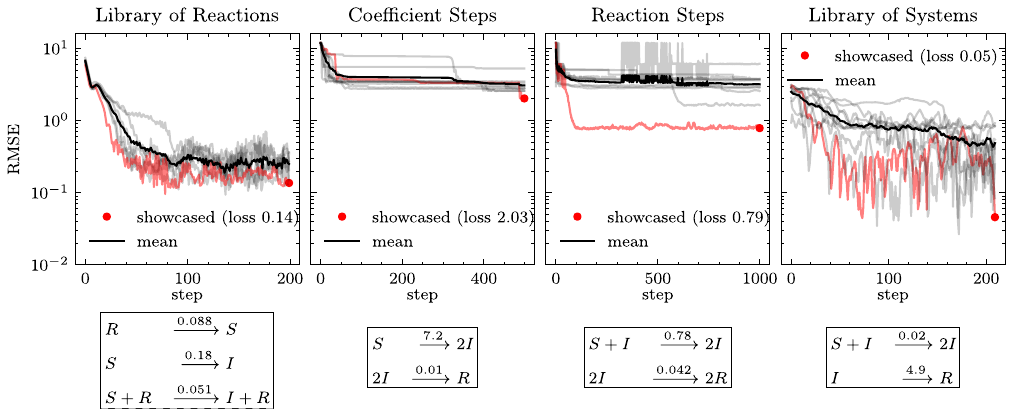}
    \vspace{-0.4cm}
    \caption{Convergence of gradient descent on the four problems (top) and chosen inferred models (bottom). Progress on the unsmoothed objective, the optimal solution has a loss of about $0.01$ (depending on the inferred system's stochasticity). 
    The reaction system depicted for \emph{Library of Reactions} shows only the top 3 of the 17 learned reactions above the threshold $10^{-4}$.}
    \label{fig:loss}
\end{figure*}

\subsection{Reparametrization}
\label{sec:reparametrization}

In both parameter estimation and structure identification, order-of-magnitude differences in the rate constants pose a problem for optimization:
the appropriate step size depends on the dimension of $\mathbf{r}$.
This has been tackled in \cite{Klimovskaia2016sparseR} by a separate optimization to determine the orders of magnitude.
The authors of \cite{Nobile2022shaping} use hand-crafted and learned dilation functions.
Here, we use a simple logarithmic reparametrization, decreasing the dynamic range of the parameters:
$$\mathbf{r}' = \exp(a\mathbf{r}+c)-\exp(c)\text{, with }a=\frac{1}{4}\text{ and }c=-20$$ 

\noindent Optimizing in this space means that a step in $\mathbf{r}$ between, e.g., $0.1$ and $0.2$ is the same as between $1$ and $2$.
The specific shifting and scaling ensure (1) that the value $r_i=0$ is mapped to $0$ and (2) that the values between $10^{-4}$ and $10^2$ are sufficiently spread. 
This way, the sensitivity of the response wrt.\ changes in $\mathbf{r}$ is decreased, aiding the optimization (cf.~\autoref{fig:reparametrization}).
Specifically, in the case of our stochastic gradient estimator, this allows setting a single smoothing factor $\sigma$ for all dimensions, which would otherwise lead to oversmoothing and occlude narrow minima.

\subsection{Evaluation Setup} 

To identify the challenges and opportunities of gradient descent in the context of a stochastic simulation-based model inference, we evaluate the convergence of our four problem formulations on recovering the SIR model as parametrized in \autoref{sec:population_based_modeling}.
Our time-series reference data is generated by simulating the model until $t=1$ and collecting state snapshots at $100$ discrete, equidistant simulation times (although we generally require neither equidistance nor completeness).
For optimization, we employ the stochastic gradient estimator introduced in \autoref{sec:stochastic_gradient_estimation} and combine it with the Adam gradient descent optimizer \cite{Kingma2014adamA}.
For each problem, we manually determined hyperparameters (sample size $n$, smoothing factor $\sigma$, and learning rate $\eta$) that achieved good results.
In the order of the problems from \autoref{sec:learning_reaction_systems_with_gradient_descent}, these are $(100,0.2,1)$, $(1000,1,1)$, $(100,0.2,0.1)$, and $(100,0.2,0.5)$.
Initial parameters are drawn from problem-specific uniform distributions.
Our simple demonstration aims to minimize the root mean squared error (RMSE) between the reference and the simulation mean time-series, the latter being determined from $20$ replications.
Note that it is easily possible to change this objective, e.g., to minimizing Wasserstein distances on distribution estimates \cite{Ocal2020paramete}.
We repeat the optimization process $10$ times to account for the stochasticity.

\section{Results and Discussion}
\label{sec:discussion_of_results}

The evaluation results in \autoref{fig:loss} show the mean convergence behavior over gradient descent steps on each problem, as well as the final model inferred by a chosen optimization run. 
For the Brute Force problem, the lowest RMSE of all structures is shown.

The \emph{Library of Reactions} formulation yields a very precise fit to the input data but lacks parsimony.
Convergence is attained fast, as the objective is smooth.
Here, a parsimony-encouraging initialization, such as the horseshoe prior for Bayesian regression may be beneficial \cite{jiang2022identification}, albeit introducing bias towards certain solutions.

On \emph{Coefficient Steps}, on the other hand, the smoothed gradient descent struggles to converge to a good solution.
Our further experiments showed that convergence to very good solutions is possible, but strongly depends on the initialization.
This hints at the existence of hard-to-escape local minima.

In \emph{Reaction Steps}, the smoothed gradient should be able to capture the effects of possible alternate rankings, and we observe good initial progress toward a parsimonious solution.
Still, the decoupling of rates and structure seems to be challenging to overcome.
When the ranking vector tends to a local minimum, means of escaping it by (partially) shuffling the current ranking could help to identify better solutions in other parts of the search space. 
However, in preliminary experiments of this sort, we observed inferior results.

Being completely smooth, the brute force \emph{Library of Systems} approach is similar in convergence to the Library of Reactions.
In contrast to the latter, it is able to recover the parsimonious original model.
This indicates the ability of gradient descent to optimize a vast number of reaction systems at a time.
Since the combinatorial explosion puts larger systems out of reach, the main missing piece for this approach is a goal-driven exploration of structures.

Our initial results demonstrate a tradeoff between parsimony, goodness of fit, and scalability.
This is the result of different response surfaces and their amenability to gradient descent.
In all cases, the scaling of rate constants poses a problem, which can be dealt with by reparametrization (cf. \autoref{sec:reparametrization}). 
Whereas the rate constant space clearly places solutions of similar quality close to each other (cf. \autoref{fig:reparametrization}), it is generally unclear which steps in the structure dimension (on the coefficients in $\mathbf{C}$) lead to lower loss.
The simultaneous adjustment of both $\mathbf{C}$ and $\mathbf{r}$ further complicates solutions that try to (smoothly) enforce a certain model size.
A major step towards better convergence would thus be a combined reparametrization of $\mathbf{C}$ and $\mathbf{r}$ that enables a goal-driven exploration of structures.
Clearly, such a reparametrization must be approximate, and its existence is unclear, demanding further investigation.
Promisingly, in the related case of learning (imperative) programs, first steps have been taken in this direction \cite{matt2017grammar}. 
Besides parsimony, identifyability could be facilitated by constraining solutions on background knowledge, as for example derived from a conceptual model in a simulation study.

Beyond considering the challenges outlined above, future work may explore the application of other smooth gradient estimation schemes based on automatic differentiation, such as StochasticAD \cite{Arya2022automati} or DiscoGrad \cite{Kreikemeyer2023smoothin}.
Finally, the full potential of the simulation-based approach needs to be explored, e.g., by considering unmeasured variables and alternative loss functions.

\begin{anonsuppress}
\begin{acks}
JNK and AU acknowledge the funding of the \grantsponsor{dfg}{ Deutsche Forschungsgemeinschaft}{http://dx.doi.org/10.13039/501100001659} under Grant No.:~\grantnum[https://gepris.dfg.de/gepris/projekt/320435134]{dfg}{320435134};
PA is supported by Grant No.:~\grantnum[https://gepris.dfg.de/gepris/projekt/497901036]{dfg}{497901036}.
\end{acks}
\end{anonsuppress}

\bibliographystyle{ACM-Reference-Format}
\bibliography{references}

\end{document}